\pdfoutput=1

\documentclass[11pt]{article}

\usepackage[]{acl}

\usepackage{times}
\usepackage{latexsym}
\usepackage{todonotes}
\usepackage{lipsum}
\usepackage{booktabs}
\usepackage{array}
\usepackage{multirow, makecell}
\usepackage{graphicx}
\usepackage[normalem]{ulem}

\usepackage{etoolbox}
\usepackage{hyperref}
\usepackage{enumitem}
\usepackage{amsmath}
\usepackage{color}
\usepackage{tcolorbox}
\usepackage{CJK}
\usepackage{adjustbox}
\usepackage{caption}
\usepackage{subcaption}
\usepackage{algorithm}
\usepackage{algpseudocode}
\usepackage{float}
\usepackage[normalem]{ulem}
\usepackage{xspace}
\usepackage{enumitem}
\usepackage[normalem]{ulem}
\usepackage{soul}

\usepackage[T1]{fontenc}

\usepackage[utf8]{inputenc}

\usepackage{microtype}

\newcolumntype{L}[1]{>{\raggedright\let\newline\\\arraybackslash\hspace{0pt}}p{#1}}

\newcommand\methodfull{Knowledge Base Informed Negations\xspace}
\newcommand\method{\textsc{KBIN}\xspace}
\newcommand\cgentity{\textsc{ClaimGen-entity}\xspace}
\newcommand\cgbart{\textsc{ClaimGen-BART}\xspace}
\newcommand\nottt[1]{\bgroup\let\texttt\relax#1\egroup}

\title{Generating Scientific Claims for Zero-Shot Scientific Fact Checking}

\author{Dustin Wright\textsuperscript{$\natural$}\thanks{~~Work completed while an intern at AI2}\hspace{0.5cm} David Wadden\textsuperscript{$\sharp$}\hspace{0.5cm} Kyle Lo\textsuperscript{$\flat$}\hspace{0.5cm} \\ \textbf{Bailey Kuehl}\textsuperscript{$\flat$}\hspace{0.5cm} \textbf{Arman Cohan}\textsuperscript{$\flat$}\hspace{0.5cm} \textbf{Isabelle Augenstein}\textsuperscript{$\natural$}\hspace{0.5cm} \textbf{Lucy Lu Wang}\textsuperscript{$\flat$}\\
  \textsuperscript{$\natural$}Dept. of Computer Science, University of Copenhagen, Denmark \\
  \textsuperscript{$\sharp$}University of Washington, Seattle, WA, USA \\
  \textsuperscript{$\flat$}Allen Institute for Artificial Intelligence, Seattle, WA, USA \\
  \texttt{\{dw,augenstein\}@di.ku.dk}\\
  \texttt{dwadden@cs.washington.edu}\\
  \texttt{\{kylel,baileyk,armanc,lucyw\}@allenai.org}}

\begin{document}
\maketitle
\begin{abstract}
Automated scientific fact checking is difficult due to the complexity of scientific language and a lack of significant amounts of training data, as annotation requires domain expertise.
To address this challenge, 
we propose 
scientific claim generation, the task of generating one or more atomic and verifiable claims from scientific sentences, 
and demonstrate its usefulness in zero-shot fact checking for biomedical claims. 
We propose \cgbart, a new supervised method for generating claims supported by the literature, as well as \method, a novel method for generating claim negations.
Additionally, we adapt an existing unsupervised entity-centric method of claim generation to biomedical claims, which we call \cgentity. 
Experiments on zero-shot fact checking demonstrate that both \cgentity and \cgbart,
coupled with \method,
achieve up to 90\% performance of fully supervised models trained on manually annotated claims and evidence. A rigorous evaluation study demonstrates significant improvement in generated claim and negation quality over existing baselines.\footnote{Code and data available at: \href{https://github.com/allenai/scientific-claim-generation}{https://github.com/ allenai/scientific-claim-generation}}

\end{abstract}

\section{Introduction}
Scientific documents contain complex assertions about scientific processes, making it difficult to automate important tasks such as claim extraction and scientific fact checking. 
Additionally, the collection of manually annotated labels to train models on tasks with scientific data is time consuming and expensive due to the need for domain expertise~\cite{collins-etal-2017-supervised,augenstein-sogaard-2017-multi,lehman-etal-2019-inferring,DBLP:conf/emnlp/WaddenLLWZCH20,DBLP:journals/corr/abs-2104-06486}. As such, methods which require less manual annotation are especially useful in this domain. This work 
addresses this challenge by exploring how automatic generation of scientific claims can assist with dataset creation and zero-shot fact checking in the biomedical domain.

\begin{figure}[t]
  
  \centering
    \includegraphics[width=\linewidth]{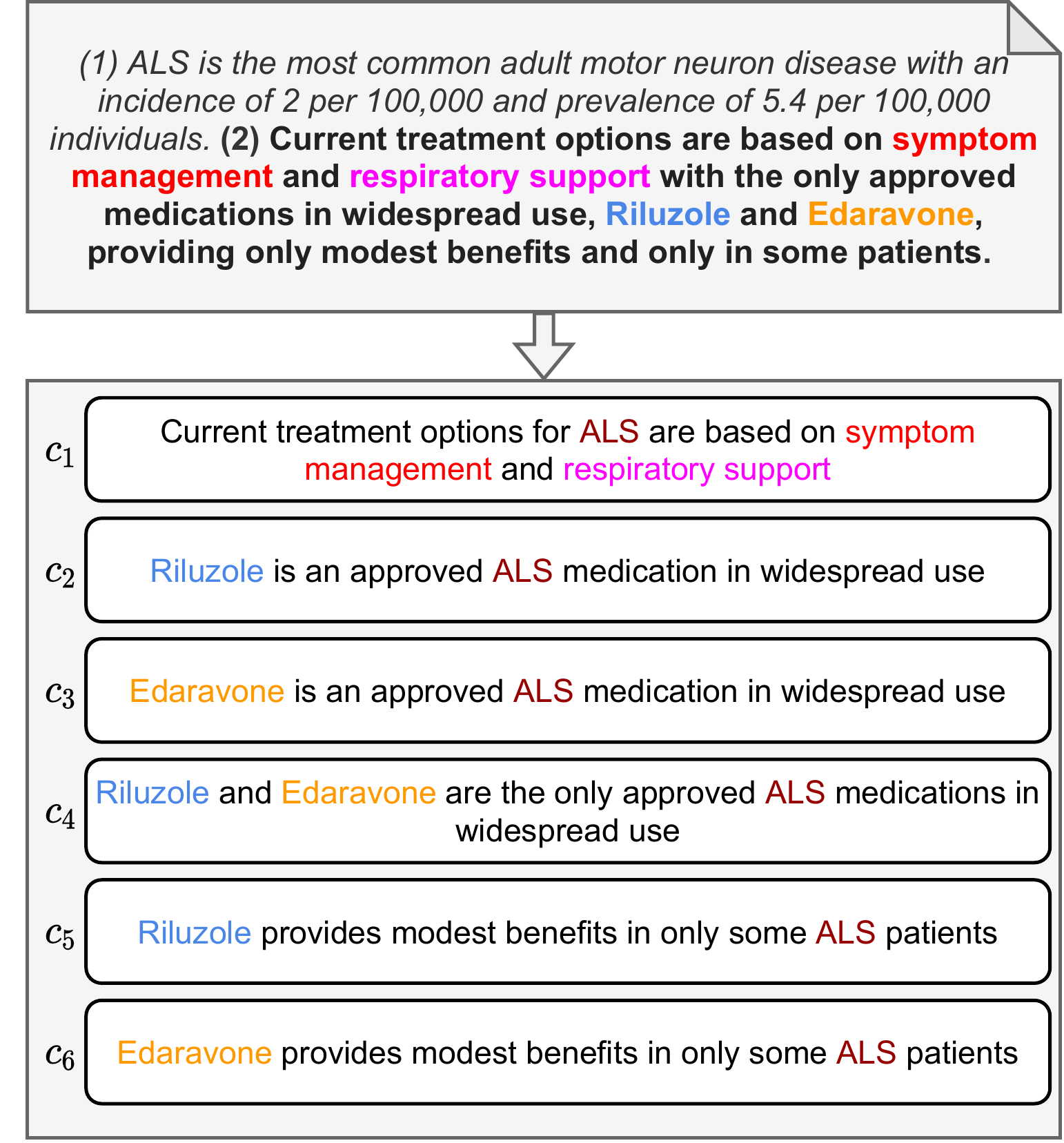}
    \caption{A complex excerpt from \citet{Mejzini2019ALSGM} (top) and the set of valid claims that can be generated from the bolded sentence (c1-c6).} 
    \label{fig:claim_generation}
\end{figure}

Being able to reduce scientific text to atomic assertions has numerous possible applications, and is known to be helpful for scientific communication and machine processing of scientific concepts~\cite{DBLP:conf/esws/KuhnBNK13}. 
Claim generation can enable zero-shot fact checking, reducing the need for expert-labeled data~\cite{DBLP:conf/acl/PanCXKW20},
and can be used to expand existing datasets such as \citet{DBLP:conf/emnlp/WaddenLLWZCH20} and \citet{DBLP:conf/acl/SaakyanCM20} without additional manual annotation. 
In this work we focus on the use of claim generation in scientific fact checking, demonstrating that claim generation enables zero-shot biomedical fact checking.

Generating scientific claims involves distilling a complex scientific sentence into one or more valid claims (see examples in \autoref{fig:claim_generation}).
As in previous work, we focus on biomedical claims as biomedical literature has long been a major focus in scientific natural language processing, as well as scientific fact checking~\cite{DBLP:conf/acl/SaakyanCM20,DBLP:conf/emnlp/WaddenLLWZCH20,kotonya-toni-2020-explainable}. While in \citet{DBLP:conf/emnlp/WaddenLLWZCH20}, claims were rewritten by domain experts from complex citation sentences (citances), we propose methods for automatically generating claims and claim negations from this source. 

Similar to other generation tasks, evaluating the quality of generated output requires multiple judgements beyond the fluency of the generated text, e.g., whether each claim is faithful to the source sentence, and is understandable on its own~\cite{DBLP:journals/corr/abs-2008-12009}. However, there are also other quality attributes that are important to assess specifically for scientific claims, such as whether each claim is atomic or check-worthy \citep{Wright2020ClaimCD}. Given this, we propose a set of manual evaluation criteria and annotation guidelines for evaluating claim generation (\S\ref{sec:quality_eval}).

Additionally, when generating claims to build datasets for tasks such as fact checking, a major challenge is creating refuted claims as negative training instances. 
Previous work has proposed automatic ways of generating refutations based on negating existing claims or creating claim variants via entity-replacement \citep{DBLP:conf/acl/PanCXKW20} and text-infilling using a pre-trained masked language model \citep{DBLP:conf/acl/SaakyanCM20}. We improve upon this by introducing \methodfull (\method), a principled method to generate refutations that performs entity-replacement using the relations and learned embeddings of entities in a domain-specific knowledge base. 

\paragraph{Contributions} In sum, our contributions are:
\begin{itemize}[noitemsep]
    \item The first study on scientific claim generation, comparing both unsupervised (\cgentity) and fully supervised (\cgbart) generation on biomedical text.
    \item \method, a novel method for generating refuted scientific claims which produces more convincing negations than previous work.
    \item Application of our claim generation methods on zero-shot scientific fact checking resulting in 90\% of the performance of a model trained on in-domain manually written claims. Additionally, a rigorous evaluation study showing that \cgbart and \method produce significantly higher quality claims and more convincing negations than previous work.
\end{itemize}


\section{Preliminaries}
\begin{figure*}[t]
  
  \centering
    \includegraphics[width=\linewidth]{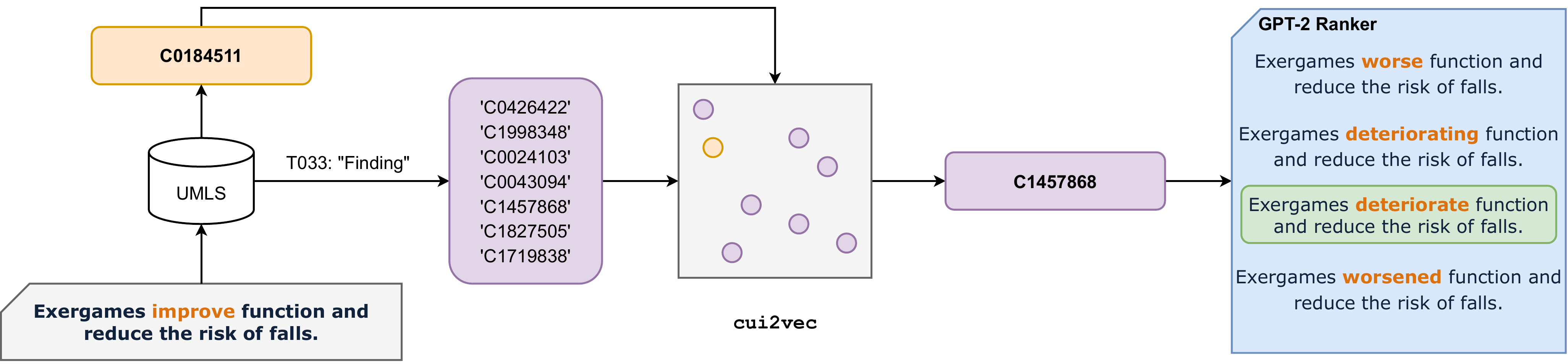}
    \caption{\method method. We start with NER and linking to UMLS using scispaCy. We then find the most similar concepts with the same type using \texttt{cui2vec}, replace the entity in the source sentence using the canonical name and aliases of similar entities, and rank them using GPT-2. Finally, from the highest ranked replacements, we select the claim which maximizes contradiction with the original claim using an external NLI model.}
    \label{fig:kbin_method}
\end{figure*}
\paragraph{Valid Claims} In this work, we define a \textit{valid claim} as one which is fluent, atomic, de-contextualized, and accurately reflects the meaning of the original sentence. Fluency is concerned with a claim being a generally well-formed English sentence, and atomicity with a claim being a ``verifiable statement expressing a finding about one aspect of a scientific entity or process,
which can be verified from a single source'' \cite{DBLP:conf/emnlp/WaddenLLWZCH20}. De-contextualilzation is concerned with a sentence being interpretable on its own, requiring none of the original surrounding text to resolve aspects of the sentence such as pronouns, abbreviations, etc., and can be handled by either directly de-contextualizing a sentence~\cite{DBLP:journals/tacl/ChoiPLKDC21} or by ensuring that all of the context sentences are available to a model~\cite{DBLP:journals/corr/abs-2112-01640}. Check-worthy claims in the wild may not be fluent, atomic, or de-contextualized, however it is useful to generate such claims as they have been shown to be useful for automated processing of science concepts~\cite{DBLP:conf/esws/KuhnBNK13} and scientific fact checking \citep{DBLP:conf/emnlp/WaddenLLWZCH20}.

\paragraph{Scientific Claim Generation} At a high level, scientific claim generation is the task of distilling one or more \textit{valid claims} from one or more sentences concerned with a scientific fact. More specifically, the task is defined as: given a scientific sentence $s$ and optionally additional context sentences $X$, generate one or more claims $c_{i} \in C$ which are valid and entailed by $s$ and $X$. In the context of fact checking, we must generate claims which are either \textit{supported} or \textit{refuted} by the literature, as well as those for which \textit{not enough information} is present to make a veracity judgement, in order that they may be paired with appropriate evidence documents to serve as training data for fact checking systems.
As such, we require methods which can take the claims in $C$ which are entailed by the source sentence and generate negations to acquire \textit{refuted} claims.

\section{Generating Supported Claims}
We experiment with two generation methods designed to produce claims which are \textit{supported} by the source sentence. The first method is an entity-centric unsupervised method adapted from \citet{DBLP:conf/acl/PanCXKW20} which requires no <sentence, claim> pairs (\cgentity). We also introduce a new method that uses BART \cite{DBLP:conf/acl/LewisLGGMLSZ20} trained on a small set of <sentence, claim> pairs to directly generate claims (\cgbart). 
For each sample $i$, we refer to the input source sentence as $s_{i}$, the context sentences as $x_{l}^{(i)} \in X_{i}$ and the output claims as $C_{i}$ consisting of $k$ claims $\{c_{1}^{(i)} \dots c_{k}^{(i)}\}$ 
Following \citet{DBLP:conf/emnlp/WaddenLLWZCH20}, we use citation sentences as unlabelled sentences for generation since these provide a natural link to an evidence document. Various components of our modeling pipelines take advantage of models pretrained on datasets for NER, NLI, QA, and fact-checking. We provide an overview of these datasets in \S\ref{sec:data_info}.

\subsection{\cgentity}
We adapt the entity-centric method presented in \citet{DBLP:conf/acl/PanCXKW20} as an unsupervised claim generation approach. This method has been tested on general domain fact checking, but has not been used for science claim generation and zero-shot scientific fact checking. In particular, we re-implement the base method used for generating supported claims and adapt it to the biomedical domain, substituting in a domain specific model for named-entity recognition. 
The method consists of the following steps for a given sample $i$:
\begin{enumerate}[noitemsep]
    \item Run named entity recognition (NER) on the input text to obtain a set of named entities $E_{i}$.
    \item For each named entity $e_{j}^{(i)}$, generate a question $q_{j}^{(i)}$ about that entity which can be answered from $s_{i}$.
    \item From $q_{j}^{(i)}$, generate the declarative form of the question to obtain claim $c_{j}^{(i)}$. 
\end{enumerate}

\paragraph{Named Entity Recognition} For NER, we employ scispaCy~\cite{DBLP:conf/bionlp/NeumannKBA19}, a spaCy\footnote{\href{https://spacy.io/}{https://spacy.io/}} pipeline for scientific NLP. The NER model is trained on the MedMentions dataset~\cite{DBLP:conf/akbc/MohanL19}, which consists of 4,392 PubMed abstracts exhaustively annotated for mentions of UMLS entities~\cite{DBLP:journals/nar/Bodenreider04}. 

\paragraph{Question Generation} For question generation, we use BART trained on questions from SQuAD~\cite{DBLP:conf/emnlp/RajpurkarZLL16}. As input for training, we encode a concatenation of the context and answer text from a given SQuAD question, and train the model to decode the question. During inference, we concatenate the source sentence $s_{i}$ and an entity $e_{j}^{(i)}$ and sample a question $q_{j}^{(i)}$ for this pair using beam search.

\paragraph{Question to Claim} Finally, as in \citet{DBLP:conf/acl/PanCXKW20}, we use a second BART model to generate declarative claims from questions. We train the model on the QA2D dataset~\cite{DBLP:journals/corr/abs-1809-02922}, which contains declarative full sentences paired with questions and their answer from SQuAD. The model is trained by encoding a concatenation of the question and answer, and decoding the full declarative sentence. At inference time, we concatenate and encode $q_{j}^{(i)}$ and $e_{j}^{(i)}$, and use beam search at the decoder to generate a claim $c_{j}^{(i)}$.

\subsection{\cgbart} 
We introduce a fully-supervised model for claim generation based on BART trained on <citance, claim> pairs. For this, we use the manual citance re-writes 
released by the SciFact authors,\footnote{\href{https://github.com/allenai/scifact/blob/master/doc/claims-with-citances.md}{https://github.com/allenai/scifact/blob/master/doc/claims-with-citances.md}}
which consist of citances from scientific papers rewritten as one or more atomic claims which are directly entailed by the citance. 

For training, we encode the citance, as well as the sentences immediately before and after the citance (the context), and train the decoder to generate claims directly. We choose to encode the context as well to help \textit{de-contextualize} generated claims. We concatenate the citance and context using a double pipe (i.e. $X_{i}$\texttt{||}$s_{i}$), and train the encoder to generate one claim at a time. 
We use top-$k$ sampling to generate multiple claims, with $k$ set to the number of noun chunks in the original source citance.\footnote{We use scispaCy to identify noun chunks}

\section{Knowledge Base Informed Negations}
\label{sec:kbin}
\begin{algorithm}[t]
\caption{KBIN algorithm}
\label{alg:kbin}
\begin{algorithmic}[1]
\Function{GetNegation}{$c, \text{KB}, V, N$}
\State $E \gets \text{NER}(c)$
\State $\bar{C} \gets []$
\For{$e_{j} \text{ in } E$}
    \State $u_{j} \gets \text{LINK}(e_{j})$
    \State $R \gets \text{KB.siblings}(u_{j})$
    \State $\text{filter}(R, \text{KB.type}(u_{j}))$
    \State $dist \gets \text{cosdist}(V[u_{j}], V[R])$
    \For{$r \text{ in argsort}(dist)[:N]$}
        \State $A \gets \text{KB.aliases}(R[r])$ 
        \State $T \gets \text{replace}(c, e_{j}, a) \text{ for } a \text{ in } A$
        \State $\bar{C}\text{.add}(\text{rank\_perplexity}(T)[0])$
    \EndFor
\EndFor
\State\Return $\text{rank\_contradiction}(c, \bar{C})[0]$ 
\EndFunction
\end{algorithmic}
\end{algorithm}

\cgentity and \cgbart only produce claims which are entailed by the source sentence. Additionally, we are interested in producing claim variants which are directly refuted by the original sentence, as 
these negations are needed when building fact checking datasets and for training fact checking models. Work in \citet{DBLP:conf/emnlp/WaddenLLWZCH20} created these negations manually, and some work has begun to explore automatically generating these negations for scientific claims~\cite{DBLP:conf/acl/SaakyanCM20}.
To this end, we leverage the availability of 
large curated biomedical knowledge bases
to develop a principled approach to claim variant generation. In particular, we use the UMLS metathesaurus~\cite{DBLP:journals/nar/Bodenreider04}, which unifies hundreds of different ontologies in biomedicine, 
as a source of term replacements for negations.


We provide an overview of the KBIN algorithm in Algorithm \autoref{alg:kbin} and \autoref{fig:kbin_method}. KBIN works by first performing NER on an input claim $c$, obtaining entities $\{e_1,\dots,e_{n}\} \in E$. For each entity $e_{j}$ in $E$,  we link the entity to its unique concept $u_{j}$ in UMLS using the scispaCy entity linker. If the entity is linked, we select all concepts which are siblings to $u_{j}$ in the concept hierarchy, and which have the same semantic type (e.g. ``Clinical Drug''). We rank all selected concepts by their cosine distance to the entity concept using pre-trained UMLS concept vectors, retaining the top 20 closest concepts. For this, we use \texttt{cui2vec}~\cite{DBLP:conf/psb/BeamKSFWPSCK20}, which contains pre-trained concept vectors for 108,477 concepts from UMLS trained on medical documents from diverse sources.

For each of the related concepts, we generate candidate claim variants by replacing the entity text in the original claim with the canonical name and aliases of the related concept from UMLS. We rank all replacement sentences by their perplexity using a pre-trained GPT-2 model~\cite{radford2019language},
keeping the sentence with least perplexity for each replacement.
Finally, from among these most fluent sentences, we select the replacement which maximizes the NLI prediction of \textit{contradiction} with the original claim.
For this, we use a RoBERTa model~\cite{DBLP:journals/corr/abs-1907-11692} pre-trained on MNLI~\cite{DBLP:conf/naacl/WilliamsNB18}. 


\section{Experiments}
We investigate three primary research questions:
\begin{itemize}[noitemsep,leftmargin=9.5mm]
    \item[\textbf{RQ1}]{Do automatically generated claims enable zero-shot scientific fact checking?}
    \item[\textbf{RQ2}]{What is the percentage of high-quality claims generated using our methods?}
    \item[\textbf{RQ3}]{How does KBIN compare with previous work for claim negation in terms of generating contradictions?}
\end{itemize}
For \textbf{RQ1}, we use \cgentity and \cgbart generated claims to train a fact checking model, evaluating on the SciFact dataset~\cite{DBLP:conf/emnlp/WaddenLLWZCH20} and comparing to relevant baselines. To answer \textbf{RQ2} and \textbf{RQ3}, we design annotation criteria and perform manual evaluations with a group of expert annotators (details in \S\ref{sec:quality_eval}). 

\subsection{RQ1: Fact Checking Performance}
\label{sec:factcheck}

\paragraph{SciFact Task} The SciFact fact verification task consists of: given a claim $c$ and a corpus of scientific abstracts $D$, retrieve evidence abstracts from $D$, predict if the claim is \textit{supported} or \textit{refuted} by those documents or if there is \textit{not enough information (NEI)} to make a prediction, and optionally determine what the rationale sentences are that explain the prediction. Here we focus on the oracle abstract setting of the task, in which gold abstracts are provided to the model and there is no retrieval component. This setup exists in the scientific fact checking literature~\cite{DBLP:conf/acl/SaakyanCM20}, and allows us to focus on one component of the fact checking pipeline for evaluating the impacts of claim generation.

\paragraph{Creating Training Data for the Zero-shot Setting} We require a set of claim-abstract pairs for training where the abstract either supports, refutes, or does not provide evidence for the given claim.
We exploit citation relationships to generate claims paired with potential evidence, using citances from the CiteWorth dataset~\cite{DBLP:conf/acl/WrightA21} as source citances for generation.
\textit{Supports} claims are produced
by directly pairing a generated claim with the abstracts of documents cited by the source citance. For \textit{refutes} claims, we negate a generated claim using \method 
and pair it with the same abstract. 
For claims labelled \textit{NEI}, we pair the generated claim or negated claim with the abstract of the source document of the citance; the source document is related to the claim but presumably does not directly support or refute the claim given the need for a citation.

\paragraph{Experimental Setup} In our experimental setup, we use LongChecker~\cite{DBLP:journals/corr/abs-2112-01640}, a Longformer~\cite{DBLP:journals/corr/abs-2004-05150} model adapted for scientific fact checking. 
The model forms its input by concatenating a claim with its evidence abstract, inserting separator tokens between sentences, and uses a classification head to predict the veracity label from the representation of the \texttt{[CLS]} token.

We explore several different setups for our training data. As a baseline, we experiment with pre-training only on FEVER claims~\cite{DBLP:conf/naacl/ThorneVCM18}, which are general domain fact checking data based on Wikipedia. We also include an experiment where we manually tune a threshold for the prediction of \textit{NEI} on the SciFact training data, as we saw that the model tends to overpredict this label without any fine-tuning on in-domain data. We also provide an upper bound on performance by fine-tuning on the in-domain train split of SciFact. Finally, we experiment with both \cgentity and \cgbart as sources of training data generated from CiteWorth citances, pairing both with \method for negations. We note that though \cgbart requires manually re-written claims as training data for generating \textit{supports} claims, it does not use any claims paired with evidence manually labelled for veracity, thus making it zero-shot for the SciFact fact-checking task. In all cases we test on the SciFact dev split.
Hyperparameter information, including number of training instances, is given in \S\ref{sec:hyperparams}, and code and data will be released upon paper acceptance. In all cases, results are reported as macro-F1. 

\paragraph{Results} Our results on SciFact are given in \autoref{tab:scifact_eval}. With an upper bound of 77.70 F1, we see that a model fine-tuned on automatically generated claims is able to achieve within 90\% of the performance of a model trained on in-domain manually written claims. This is also invariant to the method used to generate claims, as both \cgentity and \cgbart produce similar results. Additionally, both methods provide significant gains over pre-training on FEVER only, especially when no threshold on \textit{NEI} claims is used but also when re-calibrating the model to predict \textit{NEI} less often.


\begin{table}
    \def\arraystretch{1.2}
    \centering
    \fontsize{10}{10}\selectfont
    \begin{tabular}{l c c c}
    \toprule 
    Method & P & R & F1 \\
    \midrule 
       FEVER only & $86.21$& $11.96$& $21.01$\\
       FEVER + thresh & $69.15$& $66.51$& $67.80$\\
       SciFact (Upper Bound) & $77.88$& $77.51$& $77.70$\\
       \midrule
       \cgentity & $\mathbf{72.86}$& $69.38$& $\mathbf{71.08}$ \\
       \cgbart & $64.09$& $\mathbf{79.43}$& $70.94$ \\
    \bottomrule 

    \end{tabular}
    \caption{Results for veracity prediction on the SciFact dataset using different sources of training data.} 
    \label{tab:scifact_eval}
\end{table}



\subsection{RQ2: Claim Quality Evaluation}
\label{sec:quality_eval}
\begin{table*}[t]
    \def\arraystretch{1.3}
    \centering
    \fontsize{10}{10}\selectfont
    \begin{tabular}{l p{12cm}}
    \toprule 
    Metric & Labels\\
    \midrule 
    \multirow{3}{*}{Fluency} & 3 - The claim contains no grammatical errors and its meaning can be understood \\ 
    & 2 - The claim contains some grammatical errors but is still understandable \\ 
    & 1- The claim contains many grammatical errors and cannot be understood\\
    \hline
    \multirow{3}{*}{De-Contextualized} & 1 - The claim is interpretable on its own and requires no context; the addition of the original context does not alter the meaning of the claim \\ 
    & 0 - The claim cannot be interpreted in a meaningful way without the original context\\
    \hline
    \multirow{2}{*}{Atomicity} & 1 - The claim is about a single entity/process (atomic) \\
    & 0 - The claim is non-atomic and can be broken down into multiple claims \\
    \hline
    \multirow{9}{*}{Faithfulness} & 5 - The claim is correct and fully supported and complete with respect to the original sentence and context\\ 
    & 4 - The claim is correct with respect to the original sentence and context but leaves out information from the original sentence and context\\ 
    & 3 - The claim is related to the original sentence and does not contain incorrect information but is not explicitly stated in the original sentence\\ 
    & 2 - The claim contains explicitly incorrect information relative to the original sentence and context\\ 
    & 1 - The claim has nothing to do with the original sentence\\
    \bottomrule 

    \end{tabular}
    \caption{Claim quality evaluation metrics and their possible values}
    \label{tab:quality_metrics}
\end{table*}
Next, we explore if there are differences between our methods in terms of claim quality and the percentage of valid claims. For this, we ask three expert annotators to manually assess generated claims along a number of quality criteria. One annotator has undergraduate training in the life sciences and graduate training in computer science; the other two annotators have undergraduate training in the life sciences and materials science respectively. We define a set of criteria for evaluation, given in \autoref{tab:quality_metrics}.
These criteria are inspired by the AIDA (Atomic, Independent, Declarative, and Absolute) framework for scientific claims introduced in \citet{DBLP:conf/esws/KuhnBNK13}. They are also based on similar human evaluation criteria used to assess generation quality for related tasks~\cite{DBLP:journals/corr/abs-2008-12009}. We develop an initial set of guidelines for the annotators and conduct two rounds of pilot annotations to improve instructions and increase agreement. 
For the final evaluation, we generate claims on
a set of 100 citances sampled from the CiteWorth dataset~\cite{DBLP:conf/acl/WrightA21}, which contains citations in context for over 1M citances spanning 10 domains.

\begin{table*}[t]
    \def\arraystretch{1.2}
    \centering
    \fontsize{10}{10}\selectfont
    \begin{tabular}{l c c c c c c c}
    \toprule 
    Method & Fluency & De-Con. (\%) & Atomic (\%) & Faithfulness & \# Gen & \# Accept & P\\
    \midrule 
       \cgentity & $2.51$& $55.63$& $\mathbf{85.28}$& $3.54$& $893$& $111$& $12.43$\\
       \cgbart & $\mathbf{2.74}$& $\mathbf{84.35}$& $80.65$& $\mathbf{4.15}$& $156$& $69$& $\mathbf{44.23}$ \\
       \midrule
       $\alpha$ (236 claims)& 82.74& 64.53& 58.71& 53.01& - & - & -\\
    \bottomrule 

    \end{tabular}
    \caption{Average annotation score, agreement, and claim yield for each category. De-contextualized is only annotated if fluency > 1; atomicity and faithfulness are only annotated if fluency > 1 and de-contextualized == 1. \# Gen are the total claims generated by the method, and \# Accept are the number of acceptable claims generated.}
    \label{tab:manual_eval}
\end{table*}

\begin{table*}[t]
    \def\arraystretch{1.3}
    \centering
    \fontsize{10}{10}\selectfont
    \begin{tabular}{p{8.5cm} p{4.9cm} c}
    \toprule 
    Citance & Generated & Fl,D,A,Fa\\
    \midrule 
    Due to its geographic position and geological history, the island of Sardinia is characterized by a remarkable richness of endemic species and represents one of the most prominent biodiversity hotspots in the Mediterranean basin.& The island of Sardinia is characterized by a remarkable richness of endemic species.& 3,1,1,5\\
    \midrule
    Frequently reported symptom-eliciting chemicals and environmental agents include fragranted products, motor-vehicle exhaust fumes, cleaning agents, freshly printed papers or magazines, and smoke from wood burners. &Frequently reported symptom-eliciting chemicals and environmental agents are fragranted products.& 3,1,1,5\\
    \midrule
    The herbicide inhibits EPSPS (5-enolpyruvylshikimate-3-phosphate synthase) in the shikimate pathway, which has a key role in the biosynthesis of aromatic amino acids and is required for survival of the plant.& The herbicide inhibits EPSPS in the shikimate pathway. & 3,1,1,5\\
    \midrule
    Experimental models of OA, such as the intra-articular injection of monosodium acetate (MIA), are associated with joint pathology and pain behaviour comparable to clinical OA. &OA is associated with joint pathology and pain behaviour comparable to clinical OA.& 3,1,0,4\\
    \bottomrule 

    \end{tabular}
    \caption{Sample generated claims with their ratings for (Fl)uency, (D)e-Contextualized, (A)tomicity, (Fa)ithfulness}
    \label{tab:claim_examples}
\end{table*}

We limit the citances to those from papers in biology and medicine to match the domain of 
SciFact.
Annotator agreement is measured as Krippendorff's $\alpha$ \cite{krippendorff2011computing} on 236 claims for each category except fluency, where we measure the percentage of claims where all annotators agree.\footnote{Fluency agreement is measured in terms of agreement percentage as most ratings are the same (3), thus any disagreements have an oversized influence on $\alpha$.} The annotators then assess 1,049 total claims (including the 236 shared claims). Each annotator rates all criteria for an individual claim, starting with fluency, then de-contextualized, then atomicity, then faithfulness. We are mainly interested in claim quality and yield, so annotators only annotate ``de-contextualized'' if the claim is legible (fluency > 1), and only annotate ``atomicity'' and ``faithfulness'' if the claim is also de-contextualized (so one is able to discern meaning from the claim). This results in the following rules for acceptable claims based on the definitions for the labels in each category: Fluency $>$ 1 \texttt{AND} De-Contextualized $=$ 1 \texttt{AND} Atomicity $=$ 1 \texttt{AND} Faithfulness $>$ 3. An acceptable claim is thus legible, meaningful, represents a single aspect of a scientific entity or process, and accurately reflects the information presented in the original citance.

The results of claim quality annotation are given in \autoref{tab:manual_eval}. Note that these are on claims generated by \cgentity and \cgbart (see examples in \autoref{tab:claim_examples}), and thus are only \textit{supports} claims.
We first note that inter-annotator agreement is very high for fluency and moderate across all other criteria. Generated claims are quite fluent across methods, with a small minority of instances being illegible. Unsurprisingly, \cgbart improves over \cgentity across all categories except for atomicity. This intuitively makes sense as \cgentity directly produces claims which are about a single entity. \cgentity yields a higher number of claims per citance as it generates one claim for every entity in the sentence, but the precision of acceptable claims is much lower than that of \cgbart. Thus, there is a tradeoff between the two methods between the number of claims generated and their acceptability. While higher yield could lead to higher coverage of claims in the original text, this study is left to future work.

Next, we examine the similarity between generated claims 
and manually written claims from SciFact. We generate claims for each source citance $s_{i}$ in the SciFact dev split, and calculate the ROUGE score~\cite{lin2004rouge} between each generated claim $c^{(i)}_{j}$ and each manually written claim $d^{(i)}_{k}$. From this, we take an average of the max ROUGE score for each generated claim. Formally, given $|C|$ claims we calculate:
\begin{equation*}
    score = \frac{1}{|C|}\sum_{i}\sum_{j} \max_{k}\text{ROUGE}(c^{(i)}_{j}, d^{(i)}_{k})
\end{equation*}
Our evaluation results are given in \autoref{tab:rouge_eval}.
\begin{table}
    \def\arraystretch{1.2}
    \centering
    \fontsize{10}{10}\selectfont
    \begin{tabular}{l c c c}
    \toprule 
    Method & R-1 & R-2 & R-L \\
    \midrule 
       Entity & $47.12$& $27.63$& $42.30$ \\
       BART & $\mathbf{56.58}$ & $\mathbf{40.12}$ & $\mathbf{53.38}$ \\
    \bottomrule 

    \end{tabular}
    \caption{ROUGE score between generated and manually written reference claims in the SciFact dataset} 
    \label{tab:rouge_eval}
\end{table}
Both methods produce claims which have high overlap with the reference claims, though claims generated directly using BART are significantly closer to the reference claims than those generated using \cgentity. Finally, we note the these scores are in the range of state-of-the-art models used for paraphrase generation, establishing a solid baseline for this task~\cite{DBLP:conf/emnlp/ZhouB21}.

\subsection{RQ3: Negation Evaluation}
\label{sec:negeval}
\begin{table*}
    \def\arraystretch{1.3}
    \centering
    \fontsize{10}{10}\selectfont
    \begin{tabular}{p{5cm} p{0.1cm} p{3cm} p{6.2cm}}
    \toprule 
    Original Claim & & Method & Generated Negation\\
    \midrule
    Tonic signaling from the SCFV prevents constitutive stimulation.
        & & Entity replace & Tonic signaling from the SCFV {\color{red}under care of respiratory physician (finding)} constitutive stimulation. \\ \cline{3-4}
        & & \citet{DBLP:conf/acl/SaakyanCM20} & Tonic signaling from the {\color{red}inflammatory stimulation.} \\  \cline{3-4}
        & & \method & Tonic signaling from the SCFV {\color{red}accelerates} constitutive stimulation. \\
    \midrule
    Activation of the RAC1 homolog CED-10 kills viable cells in SRGP-1 mutant \textit{Caenorhabditis Elegans}.
        & & Entity replace & Activation of the {\color{red}LASS4} homolog CED-10 kills viable cells in SRGP-1 mutant \textit{Caenorhabditis Elegans}. \\ \cline{3-4}
        & & \citet{DBLP:conf/acl/SaakyanCM20} & Activation of the RAC1 homolog CED-10 kills viable cells in SRGP-1 {\color{red}\textit{Helicobacter Elegans}}. \\ \cline{3-4}
        & & \method & Activation of the RAC1 homolog CED-10 {\color{red}mediate} viable cells in SRGP-1 mutant \textit{Caenorhabditis Elegans}. \\
    \bottomrule 

    \end{tabular}
    \caption{Example negations generated using three methods. Span replacements are highlighted in {\color{red}{red}}. In addition to replacing noun phrases, \method also has the ability to replace verb phrases as shown in these examples.}
    \label{tab:claim_negations}
\end{table*}
\begin{table}
    \def\arraystretch{1.2}
    \centering
    \fontsize{10}{10}\selectfont
    \begin{tabular}{l c c c c}
    \toprule 
     & & \multicolumn{3}{c}{Entailment} \\
    Method & Fluency & 3 & 2 & 1 \\
    \midrule 
      Entity replace & 83& 1& 81& 1\\
      \citet{DBLP:conf/acl/SaakyanCM20} & 83& 10& 64& 9\\
      \method & \textbf{93}& \textbf{15}& 75& 3\\
    \bottomrule 

    \end{tabular}
    \caption{Results for manual annotation of claim negations on 100 negations for each method. Fluent claims received annotations other than ``SKIP''.} 
    \label{tab:negation_eval}
\end{table}
Finally, we perform a manual evaluation to compare \method against other methods of negation generation. Annotators evaluate negations based on Fluency and Entailment. 
We adopt the definitions used to annotate the SNLI corpus~\cite{DBLP:conf/emnlp/BowmanAPM15}, in which the annotator is given an original claim (premise) and a generated negation (hypothesis) and asked to select from among the following options, including a SKIP option for Fluency:
\begin{itemize}[noitemsep,leftmargin=10mm,itemindent=-15pt]
    \item[]\textbf{3}\quad The hypothesis is DEFINITELY FALSE given the premise
    \item[]\textbf{2}\quad The hypothesis MIGHT BE TRUE given the premise
    \item[]\textbf{1}\quad The hypothesis is DEFINITELY TRUE given the premise
    \item[]\textbf{SKIP}~ The hypothesis contains a lot of grammatical errors and cannot be understood
\end{itemize}

We compare \method to two baselines.
The first baseline replaces a single entity in the claim with a random entity of the same type, similar to the method in \citet{DBLP:conf/acl/PanCXKW20}. The second is the proposed negation generation method in \citet{DBLP:conf/acl/SaakyanCM20}. The method is based on extracting keywords using YAKE~\cite{DBLP:journals/isci/CamposMPJNJ20} (an unsupervised method based on statistical text features), replacing those keywords using text infilling with a pre-trained language model, and selecting the replacement with the highest contradiction score using a model pre-trained for NLI. We generate negations for 100 claims using all three methods. For annotation, generated negations from all three methods are aggregated and the order of negation method randomized for each of the 100 claims.


Example negations generated by all three methods are given in \autoref{tab:claim_negations} and annotation results for fluency and entailment are given in \autoref{tab:negation_eval}. First, \method produces more fluent claims than both baselines.
Additionally, \method produces more convincing negations on average than both baselines. We observe that the most common operation performed by all three methods is to replace a noun phrase. \method has the benefit of being able to replace many entity types 
corresponding to concepts found in UMLS, 
which also include verb phrases that encode relations. Finally, \method improves over the baseline from \citet{DBLP:conf/acl/SaakyanCM20} by producing fewer claims which are directly entailed by the source claim, i.e., that maintain the original meaning and do not negate the original claim.

\subsection{Further Analysis}
\begin{figure}[t]
  
  \centering
    \includegraphics[width=\linewidth]{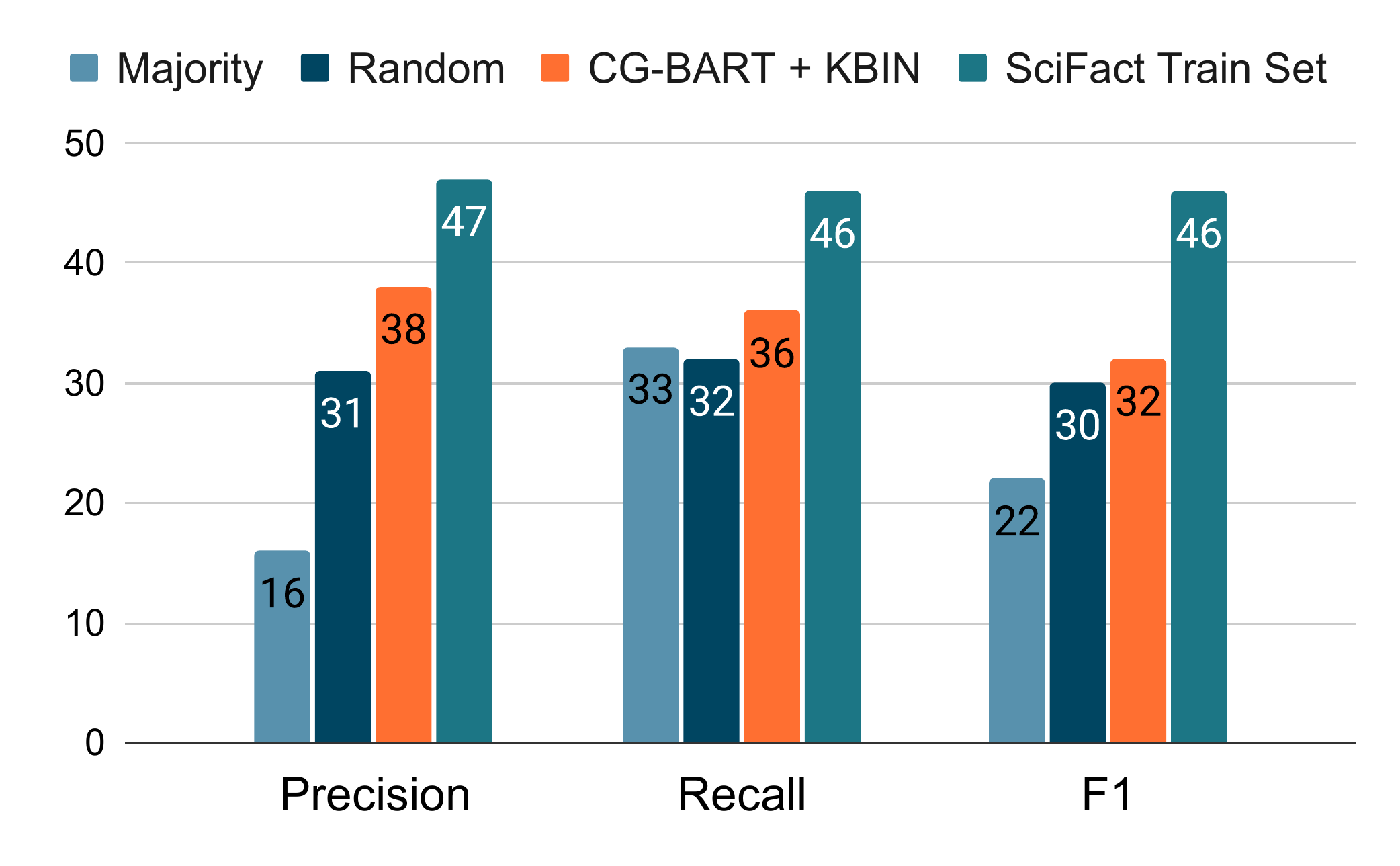}
    \caption{Fact checking performance of models trained only on claims (i.e. no evidence). Training on our generated claims result in performance closer to random (indicating fewer data artifacts) than training on the original SciFact claims.}
    \label{fig:claim_only_baseline}
\end{figure}

To give further insight into the quality of claims generated using our methods, we perform an experiment where we train and test models for scientific fact checking using claims only. This ``claim-only'' experiment helps us assess whether the negation process introduces data artifacts that can be leveraged by the model to predict veracity. We present results from training on claims generated using \cgbart and \method, compared against training on the original SciFact training data (which has manually written negations), along with random and majority baselines, in \autoref{fig:claim_only_baseline}. 

We observe that there are likely some dataset artifacts in the original SciFact claims that lead to model performance well above the majority  and random baselines.\footnote{It is difficult to fully separate the contributions of data artifacts and model performance in this setting, i.e., there is no situation which guarantees *no* undesirable data artifacts. Performance ought to be better than a random baseline in this theoretical setting, due to the pretrained language model likely having had some exposure to the content of the claims during pretraining.} This phenomenon has been observed in general domain natural language inference datasets as well~\cite{DBLP:conf/starsem/PoliakNHRD18}. Training on claims generated using our methods results in performance that is much more proximal to random performance on the SciFact dev set, indicating that the label-associated bias in the original training data is not present and a possible domain shift between the original SciFact claims and our generated claims. This can further explain some of the performance gap we observe between zero-shot fact-checking and the upper bound of training on manually labeled training data (\autoref{tab:scifact_eval}).

\section{Related Work}

\paragraph{Scientific Fact Checking} Our work follows a line of recent literature on scientific fact checking~\cite{DBLP:conf/emnlp/WaddenLLWZCH20}. The goal of this task is to determine the veracity of claims related to scientific topics by retrieving appropriate documents from scientific literature, finding evidentiary sentences from those documents, and determining whether claims are supported, refuted, or there is not enough evidence to make a judgement. The task closely resembles the task of general domain fact-checking~\cite{DBLP:conf/naacl/ThorneVCM18,augenstein-etal-2019-multifc}. Well-performing systems on this task use large language models to perform neural document retrieval~\cite{DBLP:journals/corr/abs-2010-11930} or multi-task learning of rationale prediction and stance prediction~\cite{DBLP:conf/aaai/LiBP21, DBLP:journals/corr/abs-2112-01640}. 
Recent work on general domain fact checking has also introduced methods for adversarial generation of claims which are particularly difficult to fact-check~\cite{thorne-etal-2019-fever2,atanasova-etal-2020-generating}, and for performing the task without any labeled data~\cite{DBLP:conf/acl/PanCXKW20}. Our proposed methods extend zero-shot fact checking to the scientific domain, demonstrating that one can achieve 90\% of the inference performance of state-of-the-art systems without domain-specific labeled data.


\paragraph{Generating Training Data} Our work is also related to methods for the automatic generation of training data.
Generation of synthetic data has been used for multiple tasks, for example question answering~\cite{DBLP:conf/emnlp/DuanTCZ17,DBLP:conf/emnlp/RiabiSKSSS21}, knowledge-base completion~\cite{DBLP:conf/emnlp/SafaviZK21}, and fact-checking~\cite{DBLP:conf/acl/PanCXKW20}. Most similar to our setting, the COVID-Fact dataset~\cite{DBLP:conf/acl/SaakyanCM20} contains claims related to COVID-19 crawled from Reddit, and is constructed semi-automatically. Claims which are supported by evidence are extracted from Reddit and verified by human annotators, while negations of these claims are generated automatically via masked language model infilling. \method improves upon the negation method proposed in this work by leveraging in-domain structured knowledge via UMLS.

\section{Conclusion}
In this work, we propose the task of scientific claim generation, presenting \cgbart, \cgentity, and \method to perform the task.
We demonstrate that generated claims can be used to train a model for zero-shot scientific fact checking and obtain within 90\% of the performance of a model trained on human-written claims. Through a rigorous user study we demonstrate that \cgbart produces higher quality claims than \cgentity, and that \method produces more fluent and more convincing negations than previous work.
Work remains to improve claim generation quality and assess the impacts of generated claims in other domains of science, as well as how generated claims can be used in the evidence retrieval component of fact checking systems.
We hope that our methods will be used to facilitate future work 
by enabling faster creation of training datasets and improving the performance of models on the timely and important task of scientific fact checking.

\section*{Acknowledgements}
$\begin{array}{l}\includegraphics[width=1cm]{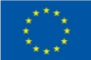} \end{array}$ This project is supported in part by the European Union's Horizon 2020 research and innovation programme under the Marie Sk\l{}odowska-Curie grant agreement No 801199, and by the United States National Science Foundation Grant OIA-2033558. 
We thank Doug Downey, Hannaneh Hajishirzi, the reviewers, and members of the Semantic Scholar research team for their valuable feedback.

\section*{Ethical Considerations}
Automated scientific fact checking has great potential value to the scientific community, as well as for addressing phenomenon such as the propagation of scientific misinformation. Our aim in releasing models for scientific claim generation is to improve the generalizability of science fact checking systems in domains with less training resources. When training our fact checking models with generated or synthetic data, there are questions regarding the veracity of the generated data and whether a model trained on inferred labels could produce trustworthy judgments. We hope that by introducing this task and models, we will enable the community to study such questions, while contributing to data curation in a domain in which such curation would normally require significant manual efforts and cost.

\bibliography{anthology,custom}

\begin{thebibliography}{41}
\expandafter\ifx\csname natexlab\endcsname\relax\def\natexlab#1{#1}\fi

\bibitem[{Atanasova et~al.(2020)Atanasova, Wright, and
  Augenstein}]{atanasova-etal-2020-generating}
Pepa Atanasova, Dustin Wright, and Isabelle Augenstein. 2020.
\newblock \href {https://doi.org/10.18653/v1/2020.emnlp-main.256} {{Generating
  Label Cohesive and Well-Formed Adversarial Claims}}.
\newblock In \emph{Proceedings of the 2020 Conference on Empirical Methods in
  Natural Language Processing (EMNLP)}, pages 3168--3177, Online. Association
  for Computational Linguistics.

\bibitem[{Augenstein et~al.(2019)Augenstein, Lioma, Wang, Chaves~Lima, Hansen,
  Hansen, and Simonsen}]{augenstein-etal-2019-multifc}
Isabelle Augenstein, Christina Lioma, Dongsheng Wang, Lucas Chaves~Lima, Casper
  Hansen, Christian Hansen, and Jakob~Grue Simonsen. 2019.
\newblock \href {https://doi.org/10.18653/v1/D19-1475} {{{M}ulti{FC}: A
  Real-World Multi-Domain Dataset for Evidence-Based Fact Checking of Claims}}.
\newblock In \emph{Proceedings of the 2019 Conference on Empirical Methods in
  Natural Language Processing and the 9th International Joint Conference on
  Natural Language Processing (EMNLP-IJCNLP)}, pages 4685--4697, Hong Kong,
  China. Association for Computational Linguistics.

\bibitem[{Augenstein and S{\o}gaard(2017)}]{augenstein-sogaard-2017-multi}
Isabelle Augenstein and Anders S{\o}gaard. 2017.
\newblock \href {https://doi.org/10.18653/v1/P17-2054} {{Multi-Task Learning of
  Keyphrase Boundary Classification}}.
\newblock In \emph{Proceedings of the 55th Annual Meeting of the Association
  for Computational Linguistics (Volume 2: Short Papers)}, pages 341--346,
  Vancouver, Canada. Association for Computational Linguistics.

\bibitem[{Beam et~al.(2020)Beam, Kompa, Schmaltz, Fried, Weber, Palmer, Shi,
  Cai, and Kohane}]{DBLP:conf/psb/BeamKSFWPSCK20}
Andrew~L. Beam, Benjamin Kompa, Allen Schmaltz, Inbar Fried, Griffin~M. Weber,
  Nathan~P. Palmer, Xu~Shi, Tianxi Cai, and Isaac~S. Kohane. 2020.
\newblock \href
  {https://psb.stanford.edu/psb-online/proceedings/psb20/Beam.pdf} {{Clinical
  Concept Embeddings Learned from Massive Sources of Multimodal Medical Data}}.
\newblock In \emph{Pacific Symposium on Biocomputing 2020, Fairmont Orchid,
  Hawaii, USA, January 3-7, 2020}, pages 295--306.

\bibitem[{Beltagy et~al.(2020)Beltagy, Peters, and
  Cohan}]{DBLP:journals/corr/abs-2004-05150}
Iz~Beltagy, Matthew~E. Peters, and Arman Cohan. 2020.
\newblock \href {http://arxiv.org/abs/2004.05150} {{Longformer: The
  Long-Document Transformer}}.
\newblock \emph{CoRR}, abs/2004.05150.

\bibitem[{Bodenreider(2004)}]{DBLP:journals/nar/Bodenreider04}
Olivier Bodenreider. 2004.
\newblock \href {https://doi.org/10.1093/nar/gkh061} {{The Unified Medical
  Language System {(UMLS):} integrating biomedical terminology}}.
\newblock \emph{Nucleic Acids Res.}, 32(Database-Issue):267--270.

\bibitem[{Bowman et~al.(2015)Bowman, Angeli, Potts, and
  Manning}]{DBLP:conf/emnlp/BowmanAPM15}
Samuel~R. Bowman, Gabor Angeli, Christopher Potts, and Christopher~D. Manning.
  2015.
\newblock \href {https://doi.org/10.18653/v1/d15-1075} {{A large annotated
  corpus for learning natural language inference}}.
\newblock In \emph{Proceedings of the 2015 Conference on Empirical Methods in
  Natural Language Processing, {EMNLP} 2015, Lisbon, Portugal, September 17-21,
  2015}, pages 632--642. The Association for Computational Linguistics.

\bibitem[{Campos et~al.(2020)Campos, Mangaravite, Pasquali, Jorge, Nunes, and
  Jatowt}]{DBLP:journals/isci/CamposMPJNJ20}
Ricardo Campos, V{\'{\i}}tor Mangaravite, Arian Pasquali, Al{\'{\i}}pio Jorge,
  C{\'{e}}lia Nunes, and Adam Jatowt. 2020.
\newblock \href {https://doi.org/10.1016/j.ins.2019.09.013} {{YAKE! Keyword
  extraction from single documents using multiple local features}}.
\newblock \emph{Inf. Sci.}, 509:257--289.

\bibitem[{Choi et~al.(2021)Choi, Palomaki, Lamm, Kwiatkowski, Das, and
  Collins}]{DBLP:journals/tacl/ChoiPLKDC21}
Eunsol Choi, Jennimaria Palomaki, Matthew Lamm, Tom Kwiatkowski, Dipanjan Das,
  and Michael Collins. 2021.
\newblock \href {https://transacl.org/ojs/index.php/tacl/article/view/2667}
  {{Decontextualization: Making Sentences Stand-Alone}}.
\newblock \emph{Trans. Assoc. Comput. Linguistics}, 9:447--461.

\bibitem[{Collins et~al.(2017)Collins, Augenstein, and
  Riedel}]{collins-etal-2017-supervised}
Ed~Collins, Isabelle Augenstein, and Sebastian Riedel. 2017.
\newblock \href {https://doi.org/10.18653/v1/K17-1021} {{A Supervised Approach
  to Extractive Summarisation of Scientific Papers}}.
\newblock In \emph{Proceedings of the 21st Conference on Computational Natural
  Language Learning ({C}o{NLL} 2017)}, pages 195--205, Vancouver, Canada.
  Association for Computational Linguistics.

\bibitem[{Demszky et~al.(2018)Demszky, Guu, and
  Liang}]{DBLP:journals/corr/abs-1809-02922}
Dorottya Demszky, Kelvin Guu, and Percy Liang. 2018.
\newblock \href {http://arxiv.org/abs/1809.02922} {{Transforming Question
  Answering Datasets Into Natural Language Inference Datasets}}.
\newblock \emph{CoRR}, abs/1809.02922.

\bibitem[{DeYoung et~al.(2021)DeYoung, Beltagy, van Zuylen, Kuehl, and
  Wang}]{DBLP:journals/corr/abs-2104-06486}
Jay DeYoung, Iz~Beltagy, Madeleine van Zuylen, Bailey Kuehl, and Lucy~Lu Wang.
  2021.
\newblock \href {http://arxiv.org/abs/2104.06486} {{MS2: Multi-Document
  Summarization of Medical Studies}}.
\newblock \emph{CoRR}, abs/2104.06486.

\bibitem[{Duan et~al.(2017)Duan, Tang, Chen, and
  Zhou}]{DBLP:conf/emnlp/DuanTCZ17}
Nan Duan, Duyu Tang, Peng Chen, and Ming Zhou. 2017.
\newblock \href {https://doi.org/10.18653/v1/d17-1090} {{Question Generation
  for Question Answering}}.
\newblock In \emph{Proceedings of the 2017 Conference on Empirical Methods in
  Natural Language Processing, {EMNLP} 2017, Copenhagen, Denmark, September
  9-11, 2017}, pages 866--874. Association for Computational Linguistics.

\bibitem[{Kotonya and Toni(2020)}]{kotonya-toni-2020-explainable}
Neema Kotonya and Francesca Toni. 2020.
\newblock \href {https://www.aclweb.org/anthology/2020.emnlp-main.623}
  {{Explainable Automated Fact-Checking for Public Health Claims}}.
\newblock In \emph{Proceedings of the 2020 Conference on Empirical Methods in
  Natural Language Processing (EMNLP)}, pages 7740--7754, Online. Association
  for Computational Linguistics.

\bibitem[{Krippendorff(2011)}]{krippendorff2011computing}
Klaus Krippendorff. 2011.
\newblock {Computing Krippendorff's alpha-reliability}.

\bibitem[{Kuhn et~al.(2013)Kuhn, Barbano, Nagy, and
  Krauthammer}]{DBLP:conf/esws/KuhnBNK13}
Tobias Kuhn, Paolo~Emilio Barbano, Mate~Levente Nagy, and Michael Krauthammer.
  2013.
\newblock \href {https://doi.org/10.1007/978-3-642-38288-8\_33} {{Broadening
  the Scope of Nanopublications}}.
\newblock In \emph{The Semantic Web: Semantics and Big Data, 10th International
  Conference, {ESWC} 2013, Montpellier, France, May 26-30, 2013. Proceedings},
  volume 7882 of \emph{Lecture Notes in Computer Science}, pages 487--501.
  Springer.

\bibitem[{Lehman et~al.(2019)Lehman, DeYoung, Barzilay, and
  Wallace}]{lehman-etal-2019-inferring}
Eric Lehman, Jay DeYoung, Regina Barzilay, and Byron~C. Wallace. 2019.
\newblock \href {https://doi.org/10.18653/v1/N19-1371} {Inferring which medical
  treatments work from reports of clinical trials}.
\newblock In \emph{Proceedings of the 2019 Conference of the North {A}merican
  Chapter of the Association for Computational Linguistics: Human Language
  Technologies, Volume 1 (Long and Short Papers)}, pages 3705--3717,
  Minneapolis, Minnesota. Association for Computational Linguistics.

\bibitem[{Lewis et~al.(2020)Lewis, Liu, Goyal, Ghazvininejad, Mohamed, Levy,
  Stoyanov, and Zettlemoyer}]{DBLP:conf/acl/LewisLGGMLSZ20}
Mike Lewis, Yinhan Liu, Naman Goyal, Marjan Ghazvininejad, Abdelrahman Mohamed,
  Omer Levy, Veselin Stoyanov, and Luke Zettlemoyer. 2020.
\newblock \href {https://doi.org/10.18653/v1/2020.acl-main.703} {{BART:
  Denoising Sequence-to-Sequence Pre-training for Natural Language Generation,
  Translation, and Comprehension}}.
\newblock In \emph{Proceedings of the 58th Annual Meeting of the Association
  for Computational Linguistics, {ACL} 2020, Online, July 5-10, 2020}, pages
  7871--7880. Association for Computational Linguistics.

\bibitem[{Li et~al.(2021)Li, Burns, and Peng}]{DBLP:conf/aaai/LiBP21}
Xiangci Li, Gully~A. Burns, and Nanyun Peng. 2021.
\newblock \href {http://ceur-ws.org/Vol-2831/paper8.pdf} {{A Paragraph-level
  Multi-task Learning Model for Scientific Fact-Verification}}.
\newblock In \emph{Proceedings of the Workshop on Scientific Document
  Understanding co-located with 35th {AAAI} Conference on Artificial
  Inteligence, SDU@AAAI 2021, Virtual Event, February 9, 2021}, volume 2831 of
  \emph{{CEUR} Workshop Proceedings}. CEUR-WS.org.

\bibitem[{Lin(2004)}]{lin2004rouge}
Chin-Yew Lin. 2004.
\newblock {ROUGE: A package for automatic evaluation of summaries}.
\newblock In \emph{Text summarization branches out}, pages 74--81.

\bibitem[{Liu et~al.(2019)Liu, Ott, Goyal, Du, Joshi, Chen, Levy, Lewis,
  Zettlemoyer, and Stoyanov}]{DBLP:journals/corr/abs-1907-11692}
Yinhan Liu, Myle Ott, Naman Goyal, Jingfei Du, Mandar Joshi, Danqi Chen, Omer
  Levy, Mike Lewis, Luke Zettlemoyer, and Veselin Stoyanov. 2019.
\newblock \href {http://arxiv.org/abs/1907.11692} {{RoBERTa: A Robustly
  Optimized BERT Pretraining Approach}}.
\newblock \emph{CoRR}, abs/1907.11692.

\bibitem[{Mejzini et~al.(2019)Mejzini, Flynn, Pitout, Fletcher, Wilton, and
  Akkari}]{Mejzini2019ALSGM}
Rita Mejzini, Loren~L Flynn, Ianthe~L. Pitout, Sue Fletcher, Steve~D. Wilton,
  and Patrick~A. Akkari. 2019.
\newblock {ALS Genetics, Mechanisms, and Therapeutics: Where Are We Now?}
\newblock \emph{Frontiers in Neuroscience}, 13.

\bibitem[{Mohan and Li(2019)}]{DBLP:conf/akbc/MohanL19}
Sunil Mohan and Donghui Li. 2019.
\newblock \href {https://doi.org/10.24432/C5G59C} {{MedMentions: A Large
  Biomedical Corpus Annotated with UMLS Concepts}}.
\newblock In \emph{1st Conference on Automated Knowledge Base Construction,
  {AKBC} 2019, Amherst, MA, USA, May 20-22, 2019}.

\bibitem[{Neumann et~al.(2019)Neumann, King, Beltagy, and
  Ammar}]{DBLP:conf/bionlp/NeumannKBA19}
Mark Neumann, Daniel King, Iz~Beltagy, and Waleed Ammar. 2019.
\newblock \href {https://doi.org/10.18653/v1/w19-5034} {{ScispaCy: Fast and
  Robust Models for Biomedical Natural Language Processing}}.
\newblock In \emph{Proceedings of the 18th BioNLP Workshop and Shared Task,
  BioNLP@ACL 2019, Florence, Italy, August 1, 2019}, pages 319--327.
  Association for Computational Linguistics.

\bibitem[{Pan et~al.(2021)Pan, Chen, Xiong, Kan, and
  Wang}]{DBLP:conf/acl/PanCXKW20}
Liangming Pan, Wenhu Chen, Wenhan Xiong, Min{-}Yen Kan, and William~Yang Wang.
  2021.
\newblock \href {https://doi.org/10.18653/v1/2021.acl-short.61} {{Zero-shot
  Fact Verification by Claim Generation}}.
\newblock In \emph{Proceedings of the 59th Annual Meeting of the Association
  for Computational Linguistics and the 11th International Joint Conference on
  Natural Language Processing, {ACL/IJCNLP} 2021, (Volume 2: Short Papers),
  Virtual Event, August 1-6, 2021}, pages 476--483. Association for
  Computational Linguistics.

\bibitem[{Poliak et~al.(2018)Poliak, Naradowsky, Haldar, Rudinger, and
  Durme}]{DBLP:conf/starsem/PoliakNHRD18}
Adam Poliak, Jason Naradowsky, Aparajita Haldar, Rachel Rudinger, and
  Benjamin~Van Durme. 2018.
\newblock \href {https://doi.org/10.18653/v1/s18-2023} {{Hypothesis Only
  Baselines in Natural Language Inference}}.
\newblock In \emph{Proceedings of the Seventh Joint Conference on Lexical and
  Computational Semantics, *SEM@NAACL-HLT 2018, New Orleans, Louisiana, USA,
  June 5-6, 2018}, pages 180--191. Association for Computational Linguistics.

\bibitem[{Pradeep et~al.(2020)Pradeep, Ma, Nogueira, and
  Lin}]{DBLP:journals/corr/abs-2010-11930}
Ronak Pradeep, Xueguang Ma, Rodrigo Nogueira, and Jimmy Lin. 2020.
\newblock \href {http://arxiv.org/abs/2010.11930} {{Scientific Claim
  Verification with VERT5ERINI}}.
\newblock \emph{CoRR}, abs/2010.11930.

\bibitem[{Radford et~al.(2019)Radford, Wu, Child, Luan, Amodei, Sutskever
  et~al.}]{radford2019language}
Alec Radford, Jeffrey Wu, Rewon Child, David Luan, Dario Amodei, Ilya
  Sutskever, et~al. 2019.
\newblock {Language Models are Unsupervised Multitask Learners}.
\newblock \emph{OpenAI blog}, 1(8):9.

\bibitem[{Rajpurkar et~al.(2016)Rajpurkar, Zhang, Lopyrev, and
  Liang}]{DBLP:conf/emnlp/RajpurkarZLL16}
Pranav Rajpurkar, Jian Zhang, Konstantin Lopyrev, and Percy Liang. 2016.
\newblock \href {https://doi.org/10.18653/v1/d16-1264} {{SQuAD: 100, 000+
  Questions for Machine Comprehension of Text}}.
\newblock In \emph{Proceedings of the 2016 Conference on Empirical Methods in
  Natural Language Processing, {EMNLP} 2016, Austin, Texas, USA, November 1-4,
  2016}, pages 2383--2392. The Association for Computational Linguistics.

\bibitem[{Riabi et~al.(2021)Riabi, Scialom, Keraron, Sagot, Seddah, and
  Staiano}]{DBLP:conf/emnlp/RiabiSKSSS21}
Arij Riabi, Thomas Scialom, Rachel Keraron, Beno{\^{\i}}t Sagot, Djam{\'{e}}
  Seddah, and Jacopo Staiano. 2021.
\newblock \href {https://aclanthology.org/2021.emnlp-main.562} {{Synthetic Data
  Augmentation for Zero-Shot Cross-Lingual Question Answering}}.
\newblock In \emph{Proceedings of the 2021 Conference on Empirical Methods in
  Natural Language Processing, {EMNLP} 2021, Virtual Event / Punta Cana,
  Dominican Republic, 7-11 November, 2021}, pages 7016--7030. Association for
  Computational Linguistics.

\bibitem[{Saakyan et~al.(2021)Saakyan, Chakrabarty, and
  Muresan}]{DBLP:conf/acl/SaakyanCM20}
Arkadiy Saakyan, Tuhin Chakrabarty, and Smaranda Muresan. 2021.
\newblock \href {https://doi.org/10.18653/v1/2021.acl-long.165} {{COVID-Fact:
  Fact Extraction and Verification of Real-World Claims on {COVID-19}
  Pandemic}}.
\newblock In \emph{Proceedings of the 59th Annual Meeting of the Association
  for Computational Linguistics and the 11th International Joint Conference on
  Natural Language Processing, {ACL/IJCNLP} 2021, (Volume 1: Long Papers),
  Virtual Event, August 1-6, 2021}, pages 2116--2129. Association for
  Computational Linguistics.

\bibitem[{Safavi et~al.(2021)Safavi, Zhu, and
  Koutra}]{DBLP:conf/emnlp/SafaviZK21}
Tara Safavi, Jing Zhu, and Danai Koutra. 2021.
\newblock \href {https://aclanthology.org/2021.emnlp-main.456} {{NegatER:
  Unsupervised Discovery of Negatives in Commonsense Knowledge Bases}}.
\newblock In \emph{Proceedings of the 2021 Conference on Empirical Methods in
  Natural Language Processing, {EMNLP} 2021, Virtual Event / Punta Cana,
  Dominican Republic, 7-11 November, 2021}, pages 5633--5646. Association for
  Computational Linguistics.

\bibitem[{Sai et~al.(2020)Sai, Mohankumar, and
  Khapra}]{DBLP:journals/corr/abs-2008-12009}
Ananya~B. Sai, Akash~Kumar Mohankumar, and Mitesh~M. Khapra. 2020.
\newblock \href {http://arxiv.org/abs/2008.12009} {{A Survey of Evaluation
  Metrics Used for NLG Systems}}.
\newblock \emph{CoRR}, abs/2008.12009.

\bibitem[{Thorne et~al.(2018)Thorne, Vlachos, Christodoulopoulos, and
  Mittal}]{DBLP:conf/naacl/ThorneVCM18}
James Thorne, Andreas Vlachos, Christos Christodoulopoulos, and Arpit Mittal.
  2018.
\newblock \href {https://doi.org/10.18653/v1/n18-1074} {{FEVER: a Large-scale
  Dataset for Fact Extraction and VERification}}.
\newblock In \emph{Proceedings of the 2018 Conference of the North American
  Chapter of the Association for Computational Linguistics: Human Language
  Technologies, {NAACL-HLT} 2018, New Orleans, Louisiana, USA, June 1-6, 2018,
  Volume 1 (Long Papers)}, pages 809--819. Association for Computational
  Linguistics.

\bibitem[{Thorne et~al.(2019)Thorne, Vlachos, Cocarascu, Christodoulopoulos,
  and Mittal}]{thorne-etal-2019-fever2}
James Thorne, Andreas Vlachos, Oana Cocarascu, Christos Christodoulopoulos, and
  Arpit Mittal. 2019.
\newblock \href {https://doi.org/10.18653/v1/D19-6601} {The {FEVER}2.0 shared
  task}.
\newblock In \emph{Proceedings of the Second Workshop on Fact Extraction and
  VERification (FEVER)}, pages 1--6, Hong Kong, China. Association for
  Computational Linguistics.

\bibitem[{Wadden et~al.(2020)Wadden, Lin, Lo, Wang, van Zuylen, Cohan, and
  Hajishirzi}]{DBLP:conf/emnlp/WaddenLLWZCH20}
David Wadden, Shanchuan Lin, Kyle Lo, Lucy~Lu Wang, Madeleine van Zuylen, Arman
  Cohan, and Hannaneh Hajishirzi. 2020.
\newblock \href {https://doi.org/10.18653/v1/2020.emnlp-main.609} {{Fact or
  Fiction: Verifying Scientific Claims}}.
\newblock In \emph{Proceedings of the 2020 Conference on Empirical Methods in
  Natural Language Processing, {EMNLP} 2020, Online, November 16-20, 2020},
  pages 7534--7550. Association for Computational Linguistics.

\bibitem[{Wadden et~al.(2021)Wadden, Lo, Wang, Cohan, Beltagy, and
  Hajishirzi}]{DBLP:journals/corr/abs-2112-01640}
David Wadden, Kyle Lo, Lucy~Lu Wang, Arman Cohan, Iz~Beltagy, and Hannaneh
  Hajishirzi. 2021.
\newblock \href {http://arxiv.org/abs/2112.01640} {Longchecker: Improving
  scientific claim verification by modeling full-abstract context}.
\newblock \emph{CoRR}, abs/2112.01640.

\bibitem[{Williams et~al.(2018)Williams, Nangia, and
  Bowman}]{DBLP:conf/naacl/WilliamsNB18}
Adina Williams, Nikita Nangia, and Samuel~R. Bowman. 2018.
\newblock \href {https://doi.org/10.18653/v1/n18-1101} {{A Broad-Coverage
  Challenge Corpus for Sentence Understanding through Inference}}.
\newblock In \emph{Proceedings of the 2018 Conference of the North American
  Chapter of the Association for Computational Linguistics: Human Language
  Technologies, {NAACL-HLT} 2018, New Orleans, Louisiana, USA, June 1-6, 2018,
  Volume 1 (Long Papers)}, pages 1112--1122. Association for Computational
  Linguistics.

\bibitem[{Wright and Augenstein(2020)}]{Wright2020ClaimCD}
Dustin Wright and Isabelle Augenstein. 2020.
\newblock \href {https://doi.org/10.18653/v1/2020.findings-emnlp.43} {{Claim
  Check-Worthiness Detection as Positive Unlabelled Learning}}.
\newblock In \emph{Findings of the Association for Computational Linguistics:
  EMNLP 2020}, pages 476--488, Online. Association for Computational
  Linguistics.

\bibitem[{Wright and Augenstein(2021)}]{DBLP:conf/acl/WrightA21}
Dustin Wright and Isabelle Augenstein. 2021.
\newblock \href {https://doi.org/10.18653/v1/2021.findings-acl.157}
  {{CiteWorth: Cite-Worthiness Detection for Improved Scientific Document
  Understanding}}.
\newblock In \emph{Findings of the Association for Computational Linguistics:
  {ACL/IJCNLP} 2021, Online Event, August 1-6, 2021}, volume {ACL/IJCNLP} 2021
  of \emph{Findings of {ACL}}, pages 1796--1807. Association for Computational
  Linguistics.

\bibitem[{Zhou and Bhat(2021)}]{DBLP:conf/emnlp/ZhouB21}
Jianing Zhou and Suma Bhat. 2021.
\newblock \href {https://aclanthology.org/2021.emnlp-main.414} {{Paraphrase
  Generation: A Survey of the State of the Art}}.
\newblock In \emph{Proceedings of the 2021 Conference on Empirical Methods in
  Natural Language Processing, {EMNLP} 2021, Virtual Event / Punta Cana,
  Dominican Republic, 7-11 November, 2021}, pages 5075--5086. Association for
  Computational Linguistics.

\end{thebibliography}
\bibliographystyle{acl_natbib}

\begin{table}[th!]
    \centering
    \fontsize{10}{10}\selectfont
    \begin{tabular}{l c}
    \toprule 
     Model & \# Params\\
    \midrule
        RoBERTa & 125M \\
        BART & 140M \\
        GPT-2 & 125M \\
        Longformer-SciFact & 438M \\
    \bottomrule 

    \end{tabular}
    \caption{Model sizes.}
    \label{tab:model_sizes}
\end{table}

\appendix

\section{Reproducibility}
\subsection{Computing Infrastructure}

All experiments were run on an Amazon Web Services p3.2xlarge instance using a Tesla V100 GPU with 16GB of RAM.



\subsection{Number of Parameters per Model}
The sizes of each of the models used in this work are given in \autoref{tab:model_sizes}.



\subsection{Hyperparameters}
\label{sec:hyperparams}
\subsubsection{Fact Checking}
\paragraph{SciFact data} Learning rate: 1e-5, 5 epochs, gradient accumulation for 8 batches, 1 sample per training batch, 16-bit precision, 809 total claims.

\paragraph{FEVER threshold} We tune the NEI threshold on the training set of SciFact, testing values in the range [1e-5, 2e-5, 3e-5, 4e-5, 5e-5, 1e-4, 2e-4, 3e-4, 4e-4, 5e-4, 1e-3, 2e-3, 3e-3, 4e-3, 5e-5, 0.01, 0.12, 0.2, 0.25, 0.4, 0.5, 0.75, 0.8, 0.8, 0.99, 0.999] and find that 5e-5 produces the best result.

\paragraph{\cgbart} Learning rate: 2e-6, 5 epochs, gradient accumulation for 8 batches, 1 sample per training batch, 16-bit precision, 1,561 total training claims.

\paragraph{\cgentity} Learning rate: 4e-8, 5 epochs, gradient accumulation for 8 batches, 1 sample per training batch, 16-bit precision, 8,592 total training claims.

\subsubsection{\cgbart} Learning rate: 2e-5, 3 epochs, linear warmup for 200 steps followed by linear decay, weight decay of 0.01, batch size of 8.

\subsection{Description of Datasets}
\label{sec:data_info}
We use a variety of datasets in this study for different components of models, training, and testing. Here we provide a description of each and in which module the dataset is used.

\paragraph{SciFact}
The SciFact dataset and rewritten claims used to train \cgbart can be found at \href{https://github.com/allenai/scifact}{https://github.com/allenai/scifact}. The dataset consists of 585 original citances with rewritten claims for each of them. Each citance consists of 1-2 rewritten claims. The SciFact rewritten claims are used to train \cgbart for direct claim generation. Additionally, SciFact contains biomedical claims paired with evidence abstracts and veracity labels in \{\textit{supports}, \textit{refutes}, \textit{not enough info}\} and is split into train, dev, and test sets. We use the train set for supervised fact checking experiments, and the dev set for testing since the test set does not come with labels.

\paragraph{FEVER}
FEVER is a general domain fact checking dataset built from Wikipedia. Like SciFact, the dataset consists of claims with paired evidence documents with labels in \{\textit{supports}, \textit{refutes}, \textit{not enough info}\}. FEVER is used as pretraining data for our fact checking models for zero-shot transfer to biomedical claims. The dataset can be found here \href{https://fever.ai/resources.html}{https://fever.ai/resources.html}.

\paragraph{MedMentions}
The MedMentions dataset is a dataset of 4,392 biomedical papers annotated with mentions of UMLS entities. It is used to train the named entity recognition and normalization models used by ScispaCy, which we used for named entity recognition in \cgentity and for normalization in \method. The dataset can be found at \href{https://github.com/chanzuckerberg/MedMentions}{https://github.com/chanzuckerberg/MedMentions}

\paragraph{UMLS}
The UMLS meta-thesaurus is a large biomedical knowledge base which unifies hundreds of different ontologies in biomedicine. UMLS is used as the source knowledge base for normalization and candidate selection for \method. Additionally, it is the knowledge base used to train \texttt{cui2vec}, which is used for candidate concept selection in \method. UMLS can be found here \href{https://www.nlm.nih.gov/research/umls/index.html}{https://www.nlm.nih.gov/research/umls/index.html}.

\paragraph{SQuAD}
The SQuAD dataset can be found at: \href{https://rajpurkar.github.io/SQuAD-explorer/}{https://rajpurkar.github.io/SQuAD-explorer/}. SQuAD is used as training data for the question generation module of \cgentity. SQuAD is a question answering dataset which contains data of the form $(q, c, a)$, where $q$ is the question, $c$ is a context document, and $a$ is an answer to the question which can be found in the context. 

\paragraph{QA2D}
The QA2D dataset can be found at: \href{https://worksheets.codalab.org/worksheets/0xd4ebc52cebb84130a07cbfe81597aaf0/}{https://worksheets.codalab.org/worksheets/ 0xd4ebc52cebb84130a07cbfe81597aaf0/}. QA2D is used in the second part of the zero-shot \cgentity model to generate declarative sentences from questions. It consists of data of the form $(s, q, a)$ where $q$ is a question, $a$ is the answer to the question, and $s$ is the declarative form of the question containing the answer. 

\paragraph{MNLI}
MNLI is a crowd-sourced collection of 433k sentence pairs annotated for textual entailment. In other words, the data consists of pairs $(p, h)$, where $p$ is the premise and $h$ is the hypothesis, and labels in \{\textit{entailment}, \textit{contradiction}, \textit{neutral}\} which say if the hypothesis entails, contradicts, or is neutral towards the premise. MNLI is used to train a RoBERTa model for entailment, which is used by \method to select the best negation among a set of generated claims for a given source citance. The dataset can be found here \href{https://cims.nyu.edu/~sbowman/multinli/}{https://cims.nyu.edu/~sbowman/multinli/}



\end{document}